\documentclass[10pt,twocolumn,letterpaper]{article}

\usepackage[pagenumbers]{cvpr} 
\usepackage[dvipsnames,table,xcdraw]{xcolor}
\usepackage{multirow}
\usepackage{booktabs}
\usepackage{array}
\usepackage{tabularx}
\usepackage{graphicx}
\usepackage{pifont}
\usepackage{makecell}
\usepackage{arydshln}
\usepackage{float}
\usepackage{longtable}
\usepackage{enumitem}

\definecolor{mlb}{RGB}{173,216,230} 
\definecolor{mlo}{RGB}{255,223,186} 

\definecolor{deepgreen}{rgb}{0.3, 0.7, 0.3}
\definecolor{sred}{rgb}{0.8, 0.0, 0.0}

\definecolor{cvprblue}{rgb}{0.21,0.49,0.74}
\usepackage[pagebackref,breaklinks,colorlinks,allcolors=cvprblue]{hyperref}

\def\paperID{7589} 
\def\confName{CVPR}
\def\confYear{2026}

\title{VL4Gaze: Unleashing Vision-Language Models for Gaze Following}

\author{
    Shijing Wang$^{1,*}$, Chaoqun Cui$^{2,*}$, Yaping Huang$^{1, \dagger}$, Hyung Jin Chang$^{3}$, Yihua Cheng$^{3}$ \\
    $^{1}$Beijing Jiaotong University, \\
    $^{2}$MAIS, Institute of Automation, Chinese Academy of Sciences, \\
    $^{3}$University of Birmingham \\
    {\tt\footnotesize \{shijingwang, yphuang\}@bjtu.edu.cn, cuichaoqun2025@ia.ac.cn, \{h.j.chang, y.cheng.2\}@bham.ac.uk}
}

\begin{document}

\twocolumn[{%
\begin{@twocolumnfalse}
\maketitle
\begin{center}
  \vspace{-6mm}
  \includegraphics[width=\linewidth]{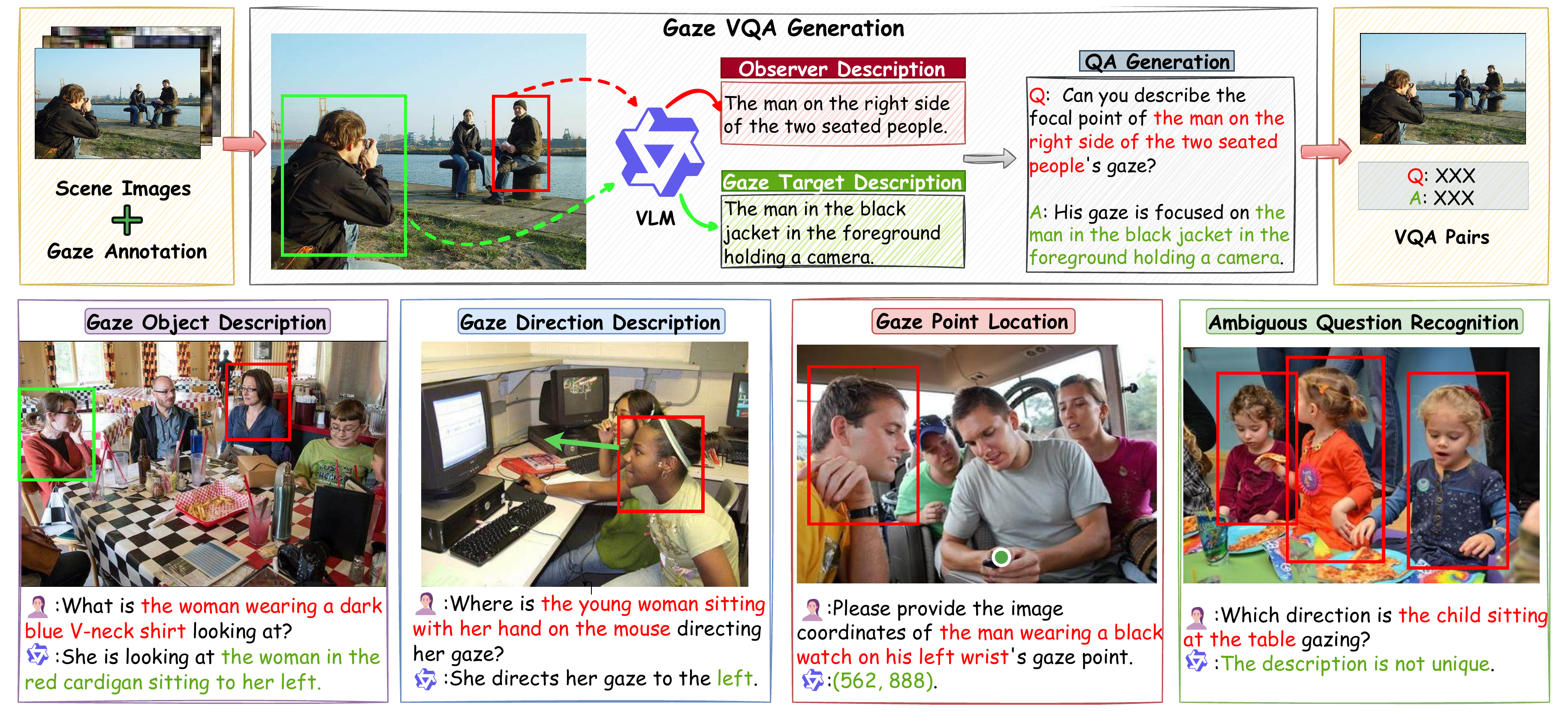}

  \vspace{-2mm}
  \captionof{figure}{We introduce the first large-scale dataset VL4Gaze to explore the capabilities of VLMs for the gaze following task. Our dataset consists of 489K text-image pairs and includes four gaze understanding tasks. Building upon this dataset, we establish the first benchmark for VLM-based gaze following, unlocking the potential of VLMs and paving the way for future research in this area.}
  \label{fig:intro}
  \vspace{3mm}
\end{center}
\end{@twocolumnfalse}%
}]

\insert\footins{\noindent\footnotesize$^*$Equal contribution. $^\dagger$Corresponding author.}

\begin{abstract}

Human gaze provides essential cues for interpreting attention, intention, and social interaction in visual scenes, yet gaze understanding remains largely unexplored in current vision-language models (VLMs). While recent VLMs achieve strong scene-level reasoning across a range of visual tasks, there exists no benchmark that systematically evaluates or trains them for gaze interpretation, leaving open the question of whether gaze understanding can emerge from general-purpose vision-language pre-training.
To address this gap, we introduce VL4Gaze, the first large-scale benchmark designed to investigate, evaluate, and unlock the potential of VLMs for gaze understanding. VL4Gaze contains 489K automatically generated question–answer pairs across 124K images and formulates gaze understanding as a unified VQA problem through four complementary tasks: (1) gaze object description, (2) gaze direction description, (3) gaze point location, and (4) ambiguous question recognition. 
We comprehensively evaluate both commercial and open-source VLMs under in-context learning and fine-tuning settings. The results show that even large-scale VLMs struggle to reliably infer gaze semantics and spatial localization without task-specific supervision. 
In contrast, training on VL4Gaze brings substantial and consistent improvements across all tasks, highlighting the importance of targeted multi-task supervision for developing gaze understanding capabilities in VLMs.
We will release the dataset and code to support further research and development in this direction.

\end{abstract}    

\begin{table}[t]
\centering
\footnotesize
\setlength{\tabcolsep}{10pt}
\renewcommand{\arraystretch}{1.1}
\caption{Comparison among traditional gaze-following datasets, recent VLM-based benchmarks for other scenes, and our first gaze-following benchmark tailored for VLMs.}
\begin{tabular}{lcc}
\toprule[1.0pt]
\textbf{Dataset} & \textbf{Annotations} & \textbf{Dataset Scale} \\
\midrule
\rowcolor[rgb]{0.902,0.902,0.902}
\multicolumn{3}{c}{\textbf{\textit{Traditional Gaze-Following Datasets}}} \\
GazeFollow~\cite{recasens2015they} & Non-text & 124K \\
VideoAttentionTarget~\cite{chong2020detecting} & Non-text & 164K \\
GOO~\cite{tomas2021goo} & Non-text & 172K \\
ChildPlay~\cite{tafasca2023childplay} & Non-text & 258K \\
\midrule
\rowcolor[rgb]{0.902,0.902,0.902}
\multicolumn{3}{c}{\textbf{\textit{Latest VLMs Benchmark (Other Scenes)}}} \\
FaceBench~\cite{wang2025facebench} & Text & 74K \\
RoadSocial~\cite{parikh2025roadsocial} & Text & 260K \\
AVQACL~\cite{wu2025avqacl} & Text & 39K \\
\midrule
\rowcolor[rgb]{0.902,0.902,0.902}
\multicolumn{3}{c}{\textbf{\textit{First Gaze-Following Benchmark for VLMs}}} \\
\textbf{VL4Gaze (Ours)} & \textbf{Text} & \textbf{489K} \\
\bottomrule[1.0pt]
\end{tabular}
\vspace{-3mm}
\label{tab:gaze_dataset_comparison}
\end{table}

\section{Introduction}

As a recent advancement, large vision-language models (VLMs) have exhibited remarkable performance across a variety of image understanding tasks, including visual grounding~\cite{NEURIPS2024_dc6319dd, zhang2024llava, rasheed2024glamm}, referring expression understanding~\cite{yang2024remamber, zhou2025dogr}, and image captioning~\cite{rotstein2024fusecap, zeng2024meacap}.
These models process natural language inputs and generate textual outputs, enabling seamless adaptation to a wide range of applications, such as embodied AI~\cite{ma2024survey, salimpour2025towards, yang2024embodied}.

Understanding human behavior is always a fundamental challenge in image understanding. Human gaze reflects attention and intention, playing a crucial role in interpreting human behavior with wide-ranging applications across multiple fields, including human-computer interaction~\cite{admoni2017social}, neuroscience~\cite{tafasca2023ai4autism}, and social psychology~\cite{capozzi2019tracking}. Conventional gaze-following tasks aim to predict where a person is looking in a frame~\cite{hu2023gfie,ryan2025gaze,chong2020detecting,tafasca2024sharingan}. Typically, these methods take a frame image and the bounding box of a specific person as input and regress the gaze target point. 
Recently, some studies~\cite{Gupta_2024_CVPR,ryan2025gaze}
have begun exploring the gaze-following task through VLMs. However, due to the absence of datasets and benchmark, these methods can only conduct preliminary explorations and lack systematic evaluation.

As illustrated in Fig.~\ref{fig:intro}, we propose VL4Gaze, the first large-scale dataset that frames gaze following within visual question answering (VQA) task, marking a new milestone for gaze following research. We introduce an automatic gaze VQA generation pipeline that converts conventional gaze-following annotations into rich textual descriptions. This pipeline leverages the image understanding capabilities of existing VLMs for text generation, significantly reducing the cost and time compared with manual annotation. To ensure reliable, natural, and diverse VQA generation, we further design a self-consistency validation mechanism and two-stage prompting strategy to enhance the overall generation quality.
As shown in Table~\ref{tab:gaze_dataset_comparison}, we build upon this pipeline to construct the VL4Gaze dataset, which consists of 489K question–answer pairs across 124K scene images, supporting both VLM fine-tuning and evaluation.

We further build the VL4Gaze benchmark based on our dataset. The benchmark defines four tasks: 1) Gaze object description, the model is required to identify the gaze target and describe it in natural language; 2) Gaze direction description, the model should provide a coarse directional answer, such as \textit{“looking to the right”}; 3) Gaze point location, the model outputs the pixel coordinates of the gaze target; 4) Ambiguous question recognition, the model should recognize when a question is ambiguous rather than hallucinating a random answer. A unified model should be trained on VL4Gaze dataset and evaluated across all four tasks. It takes as input a natural language query describing a person in the image and explicitly requesting one of the four tasks.

We evaluate VLMs on VL4Gaze to assess their gaze understanding capability. The results show that both commercial and open-source VLMs, including very large-scale ones, struggle to perform reliable gaze reasoning, \textit{even when equipped with in-context learning}. In contrast, fine-tuning on our VL4Gaze yields substantial gains across all four tasks, for example, Qwen3-VL-8B-Instruct improves gaze object description by 66\% BLEU and 29\% ROUGE-L, reduces angular error by 84\%, and decreases point-level L2 distance by 76\%, along with consistent improvements in accuracy-based metrics. Moreover, the fine-tuned model demonstrates strong cross-domain generalization on VideoAttentionTarget. These results indicate that gaze understanding does not naturally emerge from generic vision-language pre-training, and instead requires the multi-task supervision offered by our constructed VL4Gaze to become robust and transferable.

In summary, the contributions are as follows:

\begin{itemize}
\item We are the first to investigate applying VLMs to gaze understanding under a unified VQA framework.
\item We present VL4Gaze, a large-scale benchmark of diverse gaze-related question–answer pairs for evaluating and enhancing the gaze understanding ability of VLMs.
\item We fine-tune VLMs on VL4Gaze and observe consistent significant performance improvements across all gaze following-related tasks.
\end{itemize}

\section{Related Work}

\subsection{Gaze Following}
The gaze following task was introduced by Recasens et al.~\cite{recasens2015they}, aiming to predict a person’s gaze point in an image as a 2D coordinate. Their method employed a two-branch architecture, where one branch captures scene-level context and the other analyzes the cropped head to estimate gaze direction. The fused features are then used for final coordinate prediction.
Subsequent works enhanced this design by incorporating cues such as depth~\cite{fang2021dual,gupta2022modular}, body pose~\cite{gupta2022modular}, 3D head orientation~\cite{horanyi2023they}, and temporal or contextual information~\cite{chong2020detecting,gupta2024exploring}. 
More recently, Ryan et al.~\cite{ryan2025gaze} departed from the two-branch architecture and achieved strong performance using frozen vision foundation encoders with lightweight decoders, demonstrating the potential of foundation models in this task. Despite these advances, existing methods largely follow a coordinate-based input/output paradigm. To our knowledge, no prior work has explored gaze following under the visual question answering paradigm using vision-language models. 

\subsection{Vision-Language Models}
In recent years, VLMs \cite{qwen3technicalreport,grattafiori2024llama} have demonstrated strong capabilities in contextual understanding within VQA tasks. This progress is largely attributed to the breakthroughs of large language models (LLMs) \cite{bai2023qwen,touvron2023llama,jiang2024mixtral} in language modeling, as well as the strong representation ability of modern vision encoders \cite{liu2021swin,caron2021emerging,he2022masked} in visual feature extraction. However, gaze behavior, a crucial cue in human non-verbal communication, remains underexplored in current VLMs.

While VLMs possess strong scene-level understanding, which is highly relevant to gaze-following since gaze is fundamentally grounded in contextual cues, their capabilities in this task have not been systematically examined. It remains unclear whether VLMs can leverage their rich semantic reasoning to reliably interpret gaze behavior. To bridge this gap, we introduce VL4Gaze as a unified benchmark to investigate, evaluate, and unlock the potential of VLMs for gaze understanding.

\section{Automatic Gaze VQA Generation}

\begin{figure*}[ht]
    \centering
    \includegraphics[width=\linewidth]{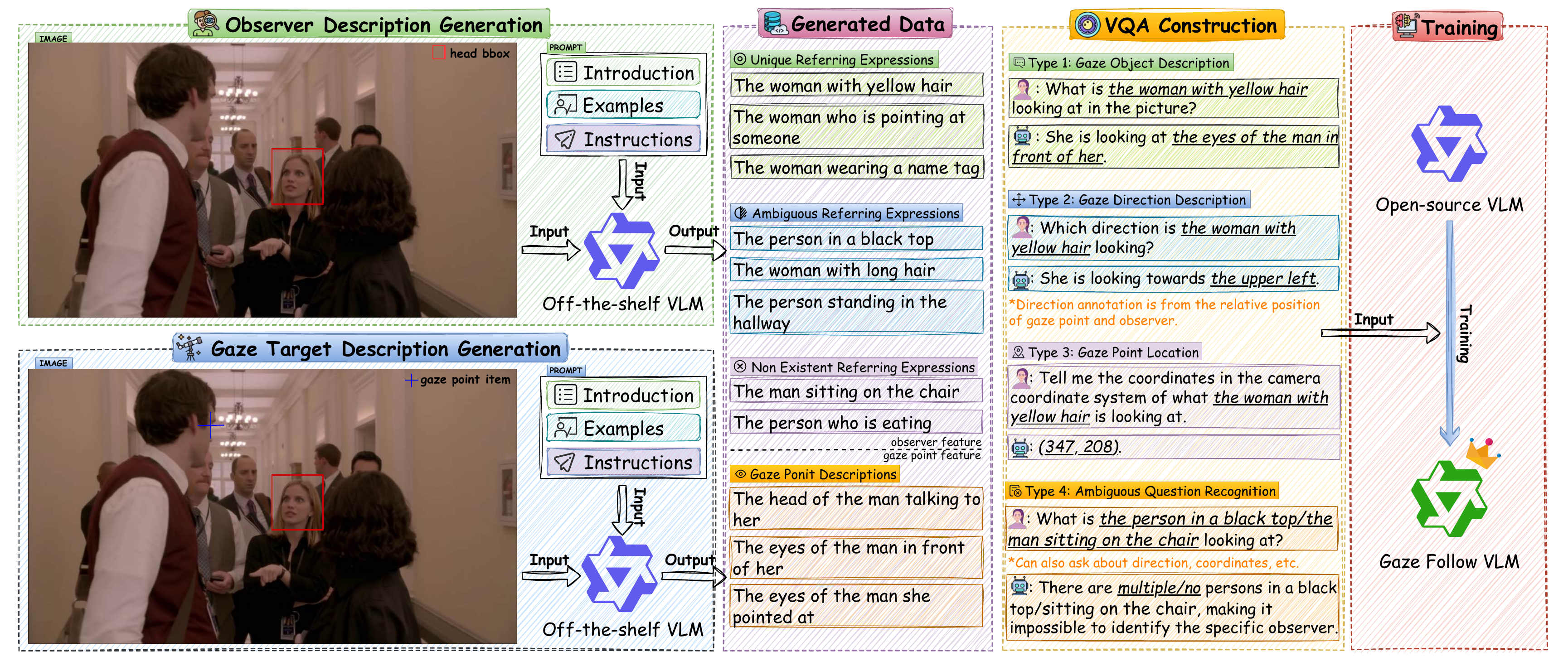}
    \caption{Overall construction pipeline of VL4Gaze benchmark.}
    \label{fig:framework}
    \vspace{-2mm}
\end{figure*}

\subsection{Overview}
Our work introduces the first large-scale benchmark for the gaze following with visual question answering (VQA) paradigm. In this section, we present an automatic Gaze VQA generation framework specifically designed for constructing the text-image annotations. Our approach leverages the image understanding capabilities of vision-language models, substantially reducing the cost and time associated with manual textual annotation. By integrating a carefully designed prompting pipeline, the proposed framework constructs reliable and diverse VQA pairs, facilitating future gaze following research based on VLMs.

As shown in Fig.~\ref{fig:framework}, our pipeline can be divided into three main stages: observer description generation, gaze target description generation, and VQA construction. We start with conventional gaze following datasets, which provide person bounding box and gaze target annotations but lack textual descriptions.
In the first stage, we generate diverse and reliable textual descriptions for observers through a self-consistency validation strategy. In the second stage, we employ a two-stage prompting strategy to produce gaze target descriptions from the observer perspective, ensuring alignment with natural answers. Finally, we construct VQA pairs by combining the generated descriptions accordingly. 

\subsection{Observer Description Generation}
The gaze behavior reflects where a person is looking in a scene. To build VQA pairs for gaze following task, We first generate textual descriptions for the observer. Specifically, given a scene image, we input the image along with the bounding box of the observer into a VLM and prompt it to generate a text description of the person (e.g., \textit{“Describe the person inside the given bounding box.”}).

However, while these prompts often produce accurate descriptions, they are not always appropriate for constructing Gaze VQA samples. For example, the VLM may correctly describe a person as \textit{“a woman wearing glasses”.} While accurate, this description becomes problematic when multiple individuals in the scene share the same characteristics, making it unclear which person the description refers to. To address this, we refine the data construction process by adjusting the prompts to guide the VLM in generating two types of descriptions: (1) unique descriptions, which unambiguously identify a single person in the scene, \ie, the observer, and (2) general descriptions, which may refer to multiple individuals. The general descriptions are particularly useful for constructing negative samples in our dataset. In other words, when given a general description, the model should response \textit{“I cannot identify the observer”} rather than hallucinating a random answer. We further divide general descriptions into two categories: existing (describing real individuals in the image) and non-existing (describing people not present in the image).

\textbf{Self-Consistency Validation.}
Trustworthiness and reliability are critical when using LLMs. While VLMs generally exhibit strong image-captioning capabilities and can generate accurate descriptions of an observer, they may not always produce unique descriptions. In our framework, we design a self-consistency validation strategy to improve the reliability of generated descriptions. First, we prompt the VLM to generate multiple candidate descriptions for the observers, including both unique and general descriptions. Next, each description is fed back into the VLM along with the same image, with the prompt: \textit{“Please tell me the number of people in the image that match this description.”} If the response for a supposed unique description is not exactly one, we discard it; the same strategy is applied to general descriptions. This self-consistency validation process effectively filters out inconsistent descriptions, ensuring high-quality annotations.

\subsection{Gaze Target Description Generation}
We also generate gaze target descriptions using VLM. Given original images and gaze points, we prompt the VLM to describe the content at the gaze location. However, we observe that such prompting cannot produce natural or consistent responses when constructing VQA pairs. For example:
\textit{“Where is the child looking?”
“She is looking at the bubble wand held by the child.”} 
While the generated text accurately describes the gaze target, humans typically answer such questions from the observer perspective, e.g., \textit{“She is looking at the the bubble wand she is holding”}. 

To improve annotation quality, we propose a two-stage prompting strategy.
We first obtain an initial description by inputting the image and gaze point into the VLM. In this stage, the VLM is prompted to focus on describing the gaze target object to ensure high-quality output. We then input the scene image, observer bounding box, gaze target point, and the initial description into VLM, prompting the VLM to rephrase this description from the observer perspective using the instruction: \textit{“Please rephrase the description from the person’s perspective.”} Since the VLM receives the initial description as input, it only performs slight modifications, reducing the risk of errors.
By explicitly separating these two stages, we obtain more natural and accurate gaze target descriptions consistent with human interpretation of gaze behavior. Notably, we generate multiple descriptions for each object. These descriptions maintain similar semantics and help promote output diversity for the model.

We also verify the gaze target annotations by prompting the VLM to determine whether the person can actually see the object. This step helps identify cases where: 1) the original gaze annotation is incorrect, and 2) the gaze target is transparent, \eg, bubbles, and the model mistakenly selects the background as the gaze target. 

\subsection{Visual Question Answering Generation}
\label{Sec: VQAG}
In the previous sections, we described the process of generating accurate and natural descriptions for both observers and gaze targets. Building upon these descriptions, we generate VQA pairs by combining observer and target descriptions with predefined question–answer templates. For each gaze behavior, one observer description and one target description are randomly sampled to form a unique VQA instance. To further enrich the dataset, we construct a template set covering four specific tasks, each with $3-4$ paraphrased templates, as detailed in the next section.

Overall, for each sample, we independently draw one observer description, one gaze target description, and one QA template from their respective pools. These components are then combined to form a complete VQA pair, ensuring diversity across the observer, target, and template dimensions. In addition, a forbidden-word list (e.g., red box) is applied to remove any descriptions containing annotation-related artifacts, ensuring that all textual content reflects genuine visual semantics.

\section{VL4Gaze Benchmark}

 \subsection{Gaze VQA Task Definition}
 \label{sec:task}
We formally define four Gaze VQA tasks and corresponding evaluation metrics. For each task, the model input contains a scene image and a textual question. Each question contains a specific person description and the desired answer category, both naturally embedded within the generated question text. The model outputs a textual description that reflects the human gaze behavior. Notably, the model should be trained on the entire dataset, including all VQA pairs across the four tasks, to develop a unified model. The performance is evaluated on these four tasks to demonstrate the generalization ability across different Gaze task.

\noindent \textbf{Task 1: Gaze Object Description (GOD).}
In this task, the model is asked to identify the object being gazed at by the observer. 
The output should be a text description of the gaze object. We also include samples where the gaze object is outside the image frame, in which case the expected answer should be similar to \textit{“The person is looking out of the frame.”}
We evaluate this task using \textbf{BLEU}~\cite{papineni2002bleu}, which measures the textual similarity between the predicted and ground-truth descriptions, and \textbf{VLM-as-a-Judge}~\cite{Gu2024SurveyLLMasAJudge}, where a VLM (specifically, Qwen3-VL-32B-Instruct~\cite{yang2025qwen3}) is used to assess the quality of the generated descriptions.

\textit{Example Q: What is the man wearing a beige sweater with a dark collar looking at in the image?
A: He is looking at the screen of the silver laptop he is using.}  

\noindent \textbf{Task 2: Gaze Direction Description (GDD).}
The model is required to identify the gaze direction of the observer. We compute the direction vector by connecting the observer and the gaze target, and discretize it into eight directions (\ie, left, right, up, down, and their four diagonal combinations). The model outputs a text description, and only the directional component of the response is evaluated. 
We use two complementary evaluation metrics. The first is classification accuracy (\textbf{ACC}), which measures the proportion of correctly predictions. The second is (\textbf{Angular Error}), where each direction is mapped to a corresponding angle, \ie, top to $0^\circ$, right to $90^\circ$, and the average angular deviation between the prediction and ground truth is computed. This provides a finer-grained assessment. 

\textit{Example Q: In which direction is the woman standing in the background holding a phone looking? A: She is looking towards the below.}

\noindent \textbf{Task 3: Gaze Point Location (GPL).}
In this task, the model is required to predict the gaze point coordinates. Unlike conventional gaze-following tasks that treat this as a numeric regression problem, we reformulate it in a VQA format, where the answer will be parsed to extract predicted gaze points. We evaluate it using the \textbf{L2} distance between the predicted and ground-truth coordinates.
In addition, traditional gaze-following benchmarks also assess whether the gaze target lies inside or outside the image. 
To maintain comparability, we compute classification accuracy (\textbf{ACC}) by prompting the model to output [-1,-1] when the gaze target is outside the frame.
Unlike the \textit{Gaze Object Description} task, which evaluates out-of-frame cases in language, this task evaluates them from a classification perspective.

\textit{Example Q: Please provide the image coordinates of the man with short dark hair and a black suit standing to the left of the man holding the microphone's gaze point. Only (x, y). If outside, (-1, -1). A: (565, 478)}

\noindent \textbf{Task 4: Ambiguous Question Recognition (AQR).}
We evaluate the model ability to recognize ambiguous questions, which is crucial for demonstrating its reliability and robustness. Ambiguity arises when the personal description in the question does not uniquely identify a single person in the image, either corresponding to multiple people or to none.
The performance is quantified using the \textbf{F1} score.

\textit{Example Q: What is the man in a black suit looking at in the image? A: The description is not unique.} 

Note that in the last three tasks, the answers do not include explicit descriptions of gaze objects. Instead, they refer to a specific predefined concept, such as \textit{left} in the gaze direction description or \textit{(-1, -1)} in the gaze point location.  For convenience, we also refer to these concepts as the gaze target description in the rest of the paper.

\begin{table}[t]
\centering
\small
\renewcommand{\arraystretch}{1.1}
\setlength{\tabcolsep}{8pt}
\caption{Comprehensive Statistics of VL4Gaze Dataset}
\vspace{-1mm}
\begin{tabular}{lcccc}
\Xhline{1.2pt}
&\multicolumn{4}{c}{\textbf{Four GazeVQA Tasks}}\\
\cline{2-5}
\textbf{VL4Gaze} & \textbf{GOD} & \textbf{GDD} & \textbf{GPL} & \textbf{AQR}\\
\midrule
\rowcolor[rgb]{0.902,0.902,0.902} \multicolumn{5}{c}{\textit{VQA Pairs (Total: 489K)}}\\\
Training & 107.5K & 121.0K & 121.0K & 121.0K\\
Test& 4.6K & 4.6K&4.6K&4.6K\\
Total&112.1K&125.6K&125.6K&125.6K\\
\midrule
\rowcolor[rgb]{0.902,0.902,0.902} \multicolumn{5}{c}{\textit{Length (words)}}\\
Question & 19.24 & 19.70 & 25.04 & 13.41 \\
Answer & 17.58 & 6.67 & 1.00 & 7.13 \\
\Xhline{1.0pt}
\vspace{-10mm}
\end{tabular}%
\label{tab:vl4gaze-stats}
\end{table}

\subsection{Dataset Statistics}

\begin{table*}[!t]
\centering

\caption{
Comparison between commercial VLM APIs and open-source VLMs on VL4Gaze. 
For commercial APIs, evaluation is conducted on a randomly sampled VL4Gaze-1K subset due to the high cost of API usage. 
We also report the performance of our fine-tuned model on the same subset for reference.
Open-source VLMs are evaluated on the full VL4Gaze test set.
We use $^{\dagger}$ to mark the method fine-tuned on VL4Gaze.
These fine-tuned models achieve substantial improvements across all tasks. Interestingly, we found that scaling up VLMs trained for general vision tasks does not necessarily yield better performance. Larger models often hallucinate, \eg, incorrectly stating that the gaze target is out of frame. In contrast, training on our VL4Gaze dataset leads to consistent and significant performance improvements.
}

\small
\resizebox{\textwidth}{!}{%
\begin{tabular}{lccccccc}
\toprule[1.2pt]
\multirow{2}{*}{\textbf{Methods}}& \multicolumn{2}{c}{\textbf{Gaze Object Description}} & \multicolumn{2}{c}{\textbf{Gaze Direction Description}} & \multicolumn{2}{c}{\textbf{Gaze Point Location}} & \textbf{Ambigu.} \\
\cline{2-8}
& \textbf{BLEU↑} & \textbf{VLM Judge↑} & \textbf{Ang. Error↓} & \textbf{ACC↑} & \textbf{L2↓} & \textbf{ACC↑} & \boldmath{\textbf{$F1$↑}} \\

\midrule
\rowcolor[rgb]{0.902,0.902,0.902} \multicolumn{8}{c}{\textit{Comparison with Commercial VLM API (VL4Gaze-1K Test Set)}}\\
Qwen-VL-Max~\cite{bai2025qwen2} & 34.01 & 0.480 & 41.22 & 0.424 & 0.377 & 0.876 & 0.650 \\
Claude4-Opus~\cite{anthropic2024claude} & 20.93 & 0.604 & 40.83 & 0.488 & 0.208 & 0.860 & 0.687 \\
Gemini-2.5-pro~\cite{team2023gemini} & 40.89 & 0.616 & 28.79 & 0.575 & 0.143 & 0.810 & 0.697 \\
GPT-5~\cite{gpt4} & 42.95 & 0.644 & 44.82 & 0.436 & 0.217 & 0.928 & 0.695 \\
\textbf{Qwen3-VL-8B-Instruct$^{\dagger}$}~\cite{yang2025qwen3} & 64.10 & 0.696 & 7.74 & 0.852 & 0.063 & 0.988 & 0.984 \\

\midrule
\rowcolor[rgb]{0.902,0.902,0.902} \multicolumn{8}{c}{\textit{Comparison with Open-Source VLM (VL4Gaze Full Test Set)}}\\

LLaVA-NeXT-72B~\cite{li2024llava} & 18.59 & 0.395 & 75.52 & 0.206 & 0.303 & 0.839 & 0.645 \\
InternVL3-78B~\cite{zhu2025internvl3} & 19.75 & 0.250 & 58.75 & 0.303 & 0.236 & 0.564 & 0.609 \\
Qwen3-VL-32B-Instruct~\cite{yang2025qwen3} & 33.98 & 0.559 & 44.23 & 0.416 & 0.254 & 0.892 & 0.727 \\
\hdashline
Qwen3-VL-2B-Instruct~\cite{yang2025qwen3} & 27.84 & 0.122 & 88.51 & 0.154 & 0.337 & 0.953 & 0.704 \\
Qwen3-VL-2B-Instruct† ~\cite{yang2025qwen3} & 
57.54\textcolor{red}{$\blacktriangle$107\%} & 
0.592\textcolor{red}{$\blacktriangle$385\%} & 
13.13\textcolor{red}{$\blacktriangledown$85\%} & 
0.790\textcolor{red}{$\blacktriangle$413\%} & 
0.091\textcolor{red}{$\blacktriangledown$73\%} & 
\textbf{0.985}\textcolor{red}{$\blacktriangle$3\%} & 
0.996\textcolor{red}{$\blacktriangle$41\%} \\

\hdashline
Qwen3-VL-4B-Instruct~\cite{yang2025qwen3}  & 28.57 & 0.472 & 49.19 & 0.372 & 0.308 & 0.963 & 0.751 \\
Qwen3-VL-4B-Instruct$^{\dagger}$ ~\cite{yang2025qwen3} & 
60.07\textcolor{red}{$\blacktriangle$110\%} & 
0.633\textcolor{red}{$\blacktriangle$34\%} & 
9.78\textcolor{red}{$\blacktriangledown$80\%} & 
0.840\textcolor{red}{$\blacktriangle$126\%} & 
0.074\textcolor{red}{$\blacktriangledown$76\%} & 
0.983\textcolor{red}{$\blacktriangle$2\%} & 
0.996\textcolor{red}{$\blacktriangle$32\%} \\

\hdashline
Qwen3-VL-8B-Instruct~\cite{yang2025qwen3} & 36.73 & 0.564 & 57.58 & 0.334 & 0.281 & 0.950 & 0.736 \\
\textbf{Qwen3-VL-8B-Instruct$^{\dagger}$ }~\cite{yang2025qwen3} & 
\textbf{61.06}\textcolor{red}{$\blacktriangle$66\%} & 
\textbf{0.637}\textcolor{red}{$\blacktriangle$13\%} & 
\textbf{9.04}\textcolor{red}{$\blacktriangledown$84\%} & 
\textbf{0.856}\textcolor{red}{$\blacktriangle$156\%} & 
\textbf{0.067}\textcolor{red}{$\blacktriangledown$76\%} & 
0.981\textcolor{red}{$\blacktriangle$3\%} & 
\textbf{0.996}\textcolor{red}{$\blacktriangle$35\%} \\
\bottomrule[1.2pt]
\end{tabular}
}
\label{tab:mainexp}
\vspace{-1mm}
\end{table*}

Our VL4Gaze is built upon the GazeFollow dataset~\cite{recasens2015they}, a carefully annotated and widely used benchmark for gaze following. The  GazeFollow dataset contains images collected from multiple sources, including SUN~\cite{xiao2010sun}, MS COCO~\cite{lin2014microsoft}, Actions 40~\cite{yao2011human}, PASCAL~\cite{everingham2010pascal}, ImageNet detection challenge~\cite{russakovsky2015imagenet} and Places~\cite{zhou2014learning} dataset, providing a large variety of scenes and contexts. We retain the original data structure and the official train/test splits of GazeFollow to ensure compatibility and fair comparison. We generate Gaze VQA annotation using Qwen3-VL-32B-Instruct~\cite{yang2025qwen3}.

The dataset statistics are summarized in Table~\ref{tab:vl4gaze-stats}. 
VL4Gaze dataset contains VQA pairs for four Gaze VQA 
tasks as defined in Section \ref{sec:task}. 
Overall, VL4Gaze contains 489K VQA pairs across 124K scene images, including approximately 470K VQA pairs
for training, and around 19K VQA pairs 
for evaluation. 
Detailed statistics for each task are also provided in Table~\ref{tab:vl4gaze-stats}. Notably, the answer length for the GPL task is 1.00, since its output corresponds to numerical coordinates.
Additionally, as described in Section \ref{Sec: VQAG}, we generate multiple candidate descriptions for both observers and gaze targets. On average, each observer has 3.15 unique descriptions and 3.94 general descriptions
, while each gaze target has 2.32 descriptions. These descriptions will also be released to facilitate future research.

\section{Experiments}

\subsection{Setup}

\textbf{Compared Methods.} We conduct a comprehensive evaluation using representative VLMs from different model families, covering a range of parameter scales and training strategies. Specifically, we include commercial VLM model including GPT-5~\cite{gpt4}, Claude4-Opus~\cite{anthropic2024claude}, Gemini-2.5-pro~\cite{team2023gemini}, and Qwen-VL-Max~\cite{bai2025qwen2}. For open-source VLMs, we consider models with various architectures and sizes, including LLaVA-NeXT-72B~\cite{li2024llava}, InternVL3-78B~\cite{zhu2025internvl3}, and the Qwen3-VL family (2B, 4B, 8B, 32B).
Since these models are not specifically trained for gaze-related tasks, we apply an \textit{in-context learning} (ICL) strategy to prompt them to produce structured outputs comparable to our fine-tuned models.

\noindent \textbf{Implementation Details.} In our experiments, we fine-tune Qwen models of different parameter scales~\cite{yang2025qwen3}, focusing on lightweight and computationally efficient variants.
All models are fine-tuned on the proposed VL4Gaze dataset and evaluated against stronger baseline methods to demonstrate the necessity and effectiveness of gaze-specific supervision for VLMs.
To increase data diversity, we perform two rounds of sampling over the training split and train the model for one epoch accordingly.
All experiments are conducted on a single machine equipped with 4 NVIDIA H100 80GB HBM3 GPUs.

\subsection{Main Results}

\begin{table*}[t]
\centering

\caption{
Parameter sensitivity and ablation analysis for Qwen3-VL models on VL4Gaze.
Increasing the number of test-time samples (Best-of-N) and enlarging the VL4Gaze training dataset scale consistently improve performance across all gaze understanding tasks.
Additionally, jointly training on multiple gaze task types yields mutual benefits across sub-tasks, while incorporating bounding boxes as auxiliary inputs further enhances localization and regression accuracy, validating the holistic multi-task design of VL4Gaze.
}
\vspace{-1mm}
\small
\resizebox{\textwidth}{!}{%
\begin{tabular}{lccccccc}
\toprule[1.2pt]
\multirow{2}{*}{\textbf{Setting / Method}} & \multicolumn{2}{c}{\textbf{Gaze Object Description}} & \multicolumn{2}{c}{\textbf{Gaze Direction Description}} & \multicolumn{2}{c}{\textbf{Gaze Point Location}} & \textbf{Ambigu.} \\
\cline{2-8}
& \textbf{BLEU↑} & \textbf{VLM-Judge↑} & \textbf{Ang. Error↓} & \textbf{ACC↑} & \textbf{L2↓} & \textbf{ACC↑} & \boldmath{\textbf{$F1$↑}} \\
\midrule

\rowcolor[rgb]{0.902,0.902,0.902} \multicolumn{8}{l}{\textbf{\textit{(A) Best-of-N Sampling}}}\\
Best-of-1 & 61.06 & 0.637 & 9.04 & 0.856 & 0.067 & 0.981 & 0.996 \\
Best-of-2 & 67.72 & 0.730 & 6.48 & 0.896 & 0.053 & 0.988 & 0.997 \\
Best-of-4 & 73.33 & 0.801 & 5.00 & 0.921 & 0.042 & 0.993 & 0.998 \\
Best-of-8 & 77.94 & 0.851 & 4.35 & 0.930 & 0.035 & 0.994 & 0.998 \\
Best-of-16 & 81.01 &0.876 & 4.05 & 0.935 & 0.030 & 0.994 & 0.998 \\
\midrule

\rowcolor[rgb]{0.902,0.902,0.902} \multicolumn{8}{l}{\textbf{\textit{(B) Training Data Scale}}}\\
Sampling 1× & 59.46 & 0.614 & 10.44 & 0.830 & 0.075 & 0.982 & 0.994 \\
Sampling 2× & 61.06 & 0.637 & 9.04 & 0.856 & 0.067 & 0.981 & 0.996 \\
Sampling 4× & 61.57 & 0.642 & 8.12 & 0.871 & 0.063 & 0.979 & 0.996 \\
\midrule

\rowcolor[rgb]{0.902,0.902,0.902} \multicolumn{8}{l}{\textbf{\textit{(C) Ablation on Question Formulation}}}\\
\textit{w/o} supervised training & 36.73 & 0.564 & 57.58 & 0.334 & 0.281 & 0.950 & 0.736 \\
Training on object description subset & 59.02 & 0.598 & 39.05 & 0.453 & 0.164 & 0.969 & 0.696 \\
Training on full set (VL4Gaze) & 61.06 & 0.637 & 9.04 & 0.856 & 0.067 & 0.981 & 0.996 \\
\hdashline
with Additional bounding box & 61.74 & 0.645 & 8.33 & 0.867 & 0.063 & 0.980 & 0.994 \\
\textit{w/o} Additional bounding box & 61.06 & 0.637 & 9.04 & 0.856 & 0.067 & 0.981 & 0.996 \\

\bottomrule[1.2pt]
\end{tabular}
}
\label{tab:param_ablation_unified}
\vspace{2mm}
\end{table*}

\begin{figure*}[t]
    \centering
    \includegraphics[width=\linewidth]{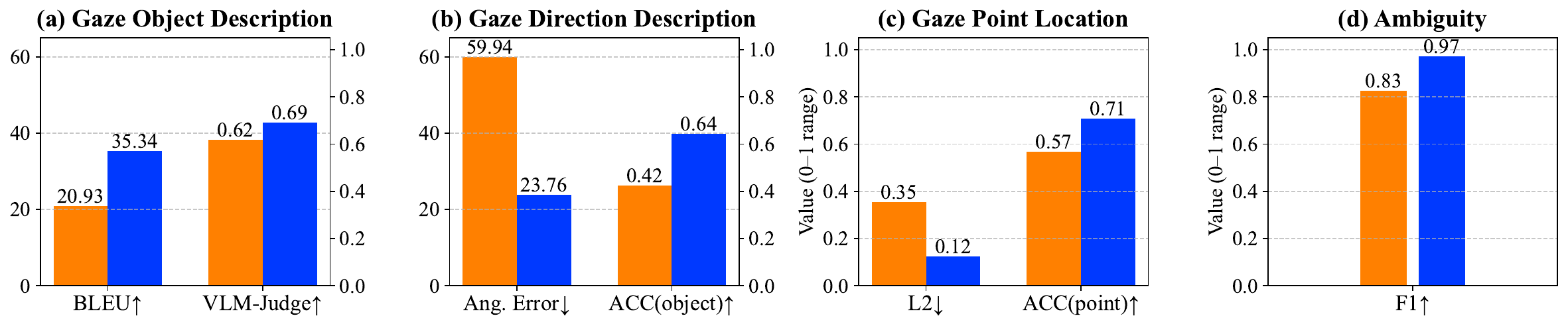}
    \vspace{-6mm}
    \caption{
    Cross-domain generalization evaluation on the VideoAttentionTarget test set.
    All models are trained exclusively on the GazeFollow portion of VL4Gaze.
    Our \textcolor{blue}{Qwen3-VL-8B-Instruct model fine-tuned on VL4Gaze} demonstrates superior gaze localization and reasoning ability under domain shift, clearly outperforming the \textcolor{orange}{non-fine-tuned Qwen3-VL-8B-Instruct} baseline.
    }
    \label{fig:crossdomain}
    \vspace{-4mm}
\end{figure*}

\begin{figure*}[t]
    \centering
    \includegraphics[width=\linewidth]{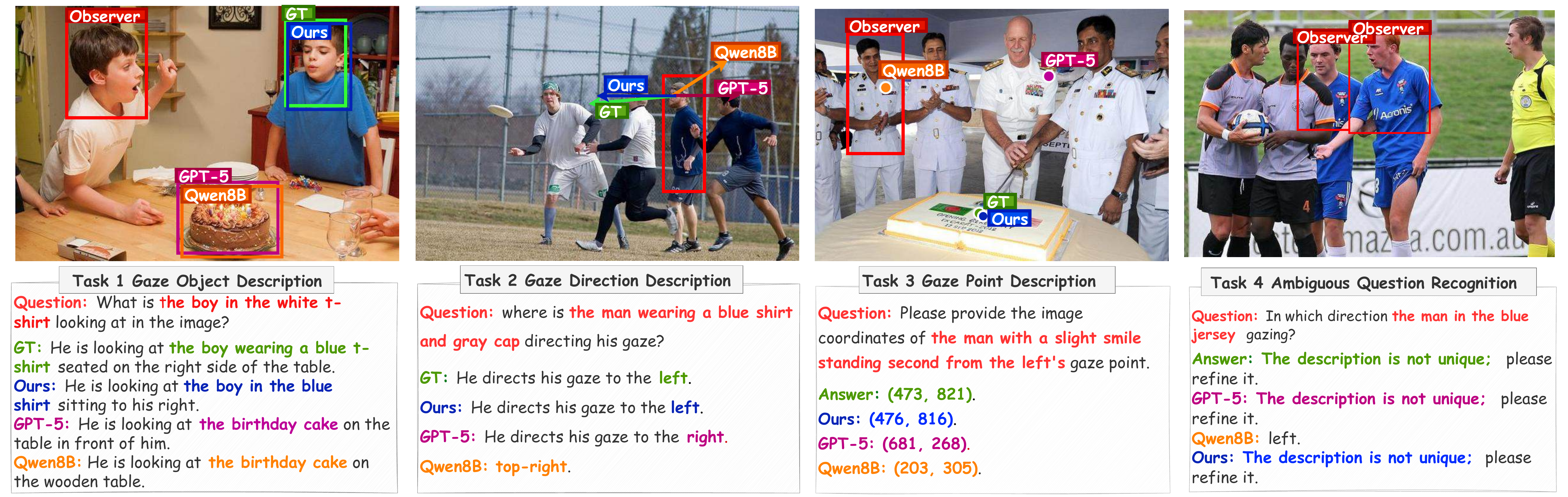}
    \vspace{-3mm}
    \caption{
    Visualization of gaze-following results from three VLMs: the commercial model \textcolor{magenta}{GPT-5}, the baseline \textcolor{orange}{Qwen3-VL-8B-Instruct}, and our fine-tuned VL4Gaze model (\textcolor{blue}{Ours}). 
    Green indicates the \textcolor{deepgreen}{ground-truth} gaze targets.
    }
    \label{fig:framework}
    \vspace{-3mm}
\end{figure*}

As summarized in Table~\ref{tab:mainexp}, we first evaluate commercial VLM APIs on the VL4Gaze-1K subset due to the high cost of API queries. Among them, GPT-5 achieves the strongest performance on gaze object description, while Gemini-2.5-Pro performs best on gaze direction estimation (lowest angular error) and gaze point localization (lowest L2). However, our fine-tuned Qwen3-VL-8B-Instruct substantially surpasses all commercial VLMs across all tasks. For example, compared with GPT-5, our model improves BLEU from 42.95 to 64.10 and VLM-as-a-judge from 0.644 to 0.696. Compared with Gemini-2.5-Pro, angular error drops from 28.79 to 7.74 and L2 decreases from 0.143 to 0.063. These results demonstrate that commercial VLMs, despite strong general reasoning capabilities, still lack robust gaze understanding without task-specific supervisions.

We then evaluate open-source VLMs using the full VL4Gaze test set. Without fine-tuning, large models such as LLaVA-NeXT-72B and InternVL3-78B perform poorly across all tasks, indicating that model scale alone does not yield effective gaze reasoning. In comparison, Qwen3-VL-32B-Instruct performs relatively better under the in-context learning setting, yet it still fall significantly short when compared to their fine-tuned counterparts.

Fine-tuning on VL4Gaze leads to consistent and substantial improvements across all Qwen model sizes. In particular, fine-tuning Qwen3-VL-8B-Instruct yields a 66\% improvement in BLEU and a 13\% improvement in VLM-as-a-judge for gaze object description, an 84\% reduction in angular error and a 156\% improvement in directional accuracy for gaze direction estimation, and a 76\% reduction in L2 distance with a 35\% improvement in F1 for gaze point localization. Furthermore, increasing model size from 2B to 4B and 8B leads to gradual and stable gains, especially on geometry-sensitive metrics such as angular error and point regression precision.

Overall, these results show that (1) fine-tuning on VL4Gaze is crucial for equipping VLMs with effective gaze reasoning capabilities, (2) scaling model capacity further enhances these abilities, and (3) ICL alone is insufficient to achieve high-quality gaze understanding. VL4Gaze thus provides an essential supervisory signal that enables VLMs to generalize robustly across diverse gaze-related tasks.

\subsection{Best-of-N Sampling}

As shown in Table~\ref{tab:param_ablation_unified}, we evaluate test-time multi-sample decoding using a Best-of-$N$ strategy. The model generates $N$ candidate responses for each query and selects the one that achieves the highest task-specific score.
This method exploits the \textit{multi-modal} nature of gaze annotation: even for the same observer and scene, different annotators may provide different yet valid gaze points~\citep{recasens2015they}.  VLMs exhibit a similar multi-modal output distribution. Sampling multiple candidates allows the model to capture this diversity and choose the most consistent prediction.

Clearly, increasing $N$ leads to clear improvements across all tasks. For Gaze Object Description, BLEU improves by 32.7\% and VLM-Judge by 37.5\%. For Gaze Direction, angular error decreases by 55.2\%, while accuracy increases by 9.2\%. For Gaze Point Localization, L2 distance decreases by 55.2\%, and accuracy increases by 1.3\%. Ambiguity detection also shows a slight gain of 0.2\%.

\subsection{Data Scale}

As shown in Table~\ref{tab:param_ablation_unified}, we analyze the effect of training data scale by varying the number of sampling passes used to construct VL4Gaze. Increasing the sampling rate from $1\times$ to $2\times$ consistently improves performance across all tasks, while scaling to $4\times$ further enhances geometric reasoning, reducing angular error (9.04→8.12) and L2 distance (0.067→0.063). These results indicate that richer linguistic and visual paraphrasing strengthens model generalization, particularly in fine-grained gaze localization.

It is worth noting that the improvement is achieved through re-sampling (see Section~\ref{Sec: VQAG}) rather than new annotation, making it both computationally and practically efficient. VL4Gaze will be released at multiple data scales to support experiments under different resource budgets.

\subsection{Ablation on Question Formulation}

As shown in Table~\ref{tab:param_ablation_unified}, removing supervised training and relying only on in-context learning leads to severe performance degradation (e.g., BLEU 36.73, angular error 57.58), underscoring the importance of VL4Gaze’s supervision.
Training only on description-type queries improves results but remains inferior to the full multi-task model (e.g., BLEU 59.02 vs.\ 61.06), indicating that gaze description, direction, and localization are mutually reinforcing.
Adding bounding boxes as additional input further reduces angular error (9.04→8.33) and L2 distance (0.067→0.063), showing that explicit spatial grounding refines both geometric and semantic reasoning.
These results highlight that supervised multi-task training and spatial grounding are key to robust gaze understanding.

\subsection{Generalization Evaluation}

To evaluate the cross-domain generalization capability of our model, we test the models trained on the GazeFollow portion of VL4Gaze on the VideoAttentionTarget dataset~\cite{chong2020detecting}, following the same QA construction process used in VL4Gaze. No additional fine-tuning is performed on the target domain. As shown in Fig.~\ref{fig:crossdomain}, in-context learning baselines achieve limited performance, especially on geometric reasoning tasks.
In contrast, our supervised fine-tuned Qwen3-VL-8B-Instruct shows strong transferability, reducing angular error by 60\% and L2 distance by 65\%, while improving directional accuracy by 52\% and gaze-point F1 by 18\%. These results show that VL4Gaze provides domain-invariant supervision that generalizes well to unseen visual environments.

\subsection{Visualization Results}

As illustrated in Fig.~\ref{fig:framework}, the visualization results clearly demonstrate the superiority of our supervised trained model compared to both the original baseline model (Qwen3-VL-8B-Instruct) and the best-performing commercial VLM (GPT-5). Although the commercial model exhibits a certain capability in gaze recognition and outperforms the baseline to some extent, it still faces challenges in capturing fine-grained gaze cues. In contrast, our fine-tuned model delivers more accurate predictions that align better with human annotations across all gaze tasks.
\section{Conclusion}

In this work, we propose VL4Gaze, the first large-scale benchmark specifically designed to investigate the capability of VLMs in gaze following.
VL4Gaze provides precise textual annotations for both observers and gaze targets, encompassing four question types: gaze point description, gaze direction description, gaze point localization, and ambiguous question recognition. This design enables the dataset to support a wide range of gaze understanding tasks.
Constructed through a scalable VLM-assisted pipeline, VL4Gaze comprises 489K QA pairs across 124K scene images. 
Experimental results show that while existing VLMs exhibit limited capability in gaze following, their performance improves substantially after training on VL4Gaze, demonstrating the benchmark’s effectiveness.

{
    \small
    \bibliographystyle{ieeenat_fullname}
    \bibliography{main}
}

\end{document}